\documentclass[11pt]{article}
\usepackage{nodalida2023}
\usepackage{times}
\usepackage{url}
\usepackage{latexsym}

\usepackage{enumitem} 
\usepackage{todonotes} 
\usepackage{hyperref} 
\usepackage{graphicx} 
\usepackage{multirow}
\usepackage{booktabs} 
\usepackage{caption}
\usepackage{subcaption}
\usepackage{enumitem} 
\usepackage{array}
\renewcommand{\arraystretch}{1.3}

\definecolor{darkblue}{rgb}{0, 0, 0.5}
\hypersetup{colorlinks=true, citecolor=darkblue, linkcolor=darkblue, urlcolor=darkblue}

\graphicspath{ {./figures/} }

\aclfinalcopy 

\title{94\% Tamrielic Proficiency: Cross-Domain Analysis \\of Fine-Grained POS Tagging on Elder Scrolls Data}
\title{Cross-Domain Evaluation of POS Taggers:\\
From Wall Street Journal to Fandom Wiki}

\author{Kia Kirstein Hansen \\
  IT University of Copenhagen \\
  {\tt kiah@itu.dk} \\\And
  Rob van der Goot \\
  IT University of Copenhagen \\
  {\tt robv@itu.dk} \\}

\date{}

\begin{document}

\maketitle
\begin{abstract}
The Wall Street Journal section of the Penn Treebank has been the de-facto standard for evaluating POS taggers for a long time, and accuracies over 97\% have been reported. However, less is known about out-of-domain tagger performance, especially with fine-grained label sets. Using data from Elder Scrolls Fandom, a wiki about the \textit{Elder Scrolls} video game universe, we create a modest dataset for qualitatively evaluating the cross-domain performance of two POS taggers: the Stanford tagger \cite{toutanova-et-al-2003} and Bilty~\cite{plank-et-al-2016}, both trained on WSJ. Our analyses show that performance on tokens seen during training is almost as good as in-domain performance, but accuracy on unknown tokens decreases from 90.37\% to 78.37\% (Stanford) and 87.84\% to 80.41\% (Bilty) across domains. Both taggers struggle with proper nouns and inconsistent capitalization.\footnote{Data, predictions and code available at:\\ \href{https://github.com/kkirsteinhansen/elder-scrolls-fandom}{github.com/kkirsteinhansen/elder-scrolls-fandom}} 
\end{abstract}

\section{Introduction}
The Wall Street Journal (WSJ) portion of the Penn Treebank~\cite{marcus-et-al-1993},  is a well-known and common benchmark for, among other tasks, POS tagging in English. The WSJ treebank consists of a collection of English newswire from the 1980s, and several existing POS taggers score over 97\% in accuracy when evaluated in-domain. In this work, we investigate how well these taggers transfer to data from wikis; online platforms where the content is continuously generated by users as a collaborative effort. Through quantitative and qualitative analyses, we explore the types of errors taggers make when tagging such out-of-\\domain data, as well as the frequencies of these errors. For this purpose, we create a new, modest English dataset to represent the wiki domain: Elder Scrolls Fandom, or ESF. Seeing as the \textit{Elder Scrolls} is a video game series, this data is expected to be topically disjoint from WSJ.

\begin{figure}
    \centering
    \includegraphics[scale=0.38]{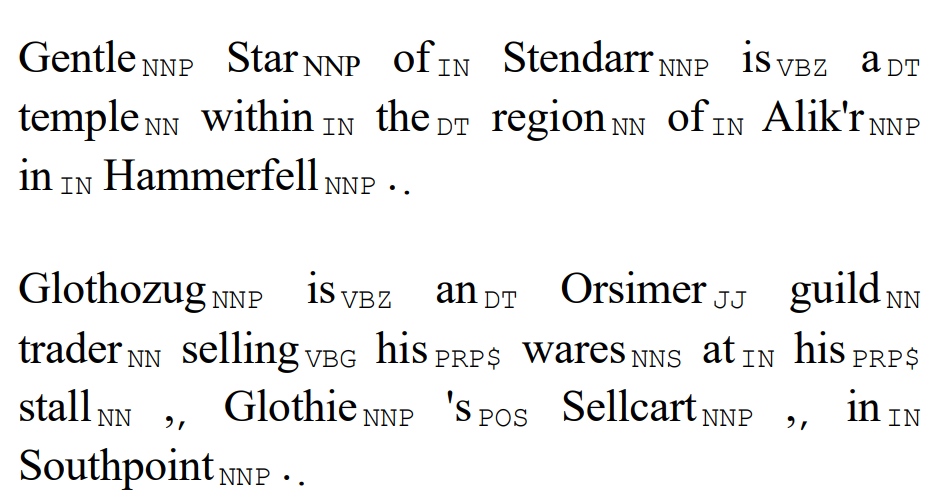}
    \caption{\label{fig:example esf sentences} Example ESF sentences with gold tags.}
\end{figure}

Cross-domain evaluation of part-of-speech taggers has for the last decade largely been focused on social media data and shallow tag sets, and POS taggers trained on newswire have been shown to suffer a decrease in accuracy on Twitter data \cite{gimpel-et-al-2011, derczynski-et-al-2013}, though this decrease is much more significant for unknown tokens than for tokens seen during training \cite{derczynski-et-al-2013}. While extensive research has been conducted on such microblog data, little is known about out-of-domain performance and errors on other domains. We therefore contribute 1) a modest evaluation data set with fine-grained POS tags on Elder Scrolls wiki data
2) a qualitative evaluation of two different types of POS taggers: traditional machine learning and deep learning 3) an in-depth analysis of challenging cases in the Fandom wiki domain.

\section{The Elder Scrolls Fandom Dataset}

\subsection{Data Collection}
In order to evaluate the newswire-trained POS taggers on out-of-domain data, we created the Elder Scrolls Fandom data set, or ESF. Fandom is a large online platform that includes wikis for video game and cinematic universes, the \textit{Elder Scrolls} being one such video game universe.
The data is user-generated content of encyclopedic nature, though it also includes text passages that are sourced directly from the video games, e.g., quest descriptions and character dialogue. Some terms describing fictional creatures, ingredients, and locations exist only in the context of these video games. The data used for ESF is English and generally well-edited, but contains some typographic errors and terms that are out-of-vocabulary for newswire. This ESF dataset thus places itself somewhere between the clean newswire data and the noisy and topically diverse microblog data.  
We used the \texttt{WikiExtractor}\footnote{\url{https://github.com/attardi/wikiextractor/}} to extract the \texttt{elderscrollsfandomcom-20200222-hi} \texttt{story.xml.7z} version of the wiki.
Target sentences were sampled at random with context (previous and next sentence) included for disambiguation. Short sentences without a clear syntactic structure were skipped\footnote{E.g., \textit{1E 1200} and \textit{2920, vol 05 - Second Seed}}. Tokenization was done using the Python \texttt{NLTK} module\footnote{\url{https://www.nltk.org/}} with manual post-correction.

\subsection{Annotation}
POS annotations were done in accordance with the guidelines for the Penn Treebank \cite{santorini-macintyre-1995}. Tokens with typographical or orthographic errors were tagged with their presumed intended meaning, e.g., the token \textit{got} in \textit{(...) the Dragonborn must got* to Filnjar (...)} was tagged as if it said \textit{go}. Two annotators annotated 20 sentences for two rounds of 10, discussing disagreements after each round and ultimately reaching a Cohen's $\kappa$ of 92.55 (agreeing on 152 out of 164 tokens), after which one annotator annotated the rest of the data for a total of 2084 tokens with 929 types. A data statement is given in Appendix~\ref{app:statement}. One notable challenge in the ESF dataset is distinguishing between proper nouns and regular nouns: Capitalization is inconsistent, and many terms are unique to the \textit{Elder Scrolls} video game universe. In some cases, even extensive context exploration did not provide enough information to make a decisive distinction; consider, e.g., 
\textit{Jazbay Grapes}. This expression is not consistently capitalized in the data, so it is unclear whether it is meant as a proper or common noun. Seeing as it refers to an alchemy ingredient and food item, we considered it semantically comparable to a fruit variety (e.g., \textit{Chardonnay}\texttt{/NNP} \textit{grape}\texttt{/NN}), and thus tagged it as \textit{Jazbay}\texttt{/NNP} (proper noun) \textit{Grapes}\texttt{/NNS} (common noun).

\begin{table}
\resizebox{1\columnwidth}{!}{%
    \begin{tabular}{ll|c c c}
        \toprule
        \normalsize{\textsc{Tgt}} & \normalsize{\textsc{Tagger}}   & \normalsize{\textsc{All}} & \normalsize{\textsc{Known}} & \normalsize{\textsc{Unknown}} \\
        \midrule
        WSJ & Stanford &\bf0.9695 & \bf0.9714 & \bf0.9037 \\
        WSJ & Bilty & 0.9677 & 0.9702 & 0.8784 \\
        \midrule
        ESF & Stanford & 0.9419 & 0.9630 & 0.7837 \\
        ESF & Bilty &  \bf0.9458 & \bf0.9647 & \bf0.8041 \\
        \bottomrule
    \end{tabular}}
    \caption{\label{tab:results} Accuracy split into known (seen during training) and unknown (unseen) tokens (12\% of ESF tokens and 2\% of WSJ dev tokens).}
\end{table}

\section{Experiments}
\subsection{POS Taggers}
We evaluated two taggers on ESF: The Stanford tagger~\cite{toutanova-et-al-2003} and Bilty~\cite{plank-et-al-2016}\footnote{We did not evaluate a transformer-based POS tagger, as 
this gave us more control over (pre-)training data and saved compute.}, both of which have reported very high accuracies on the WSJ dataset (97.24\% and 97.22\%, respectively). They also represent two different architectural implementations: The Stanford tagger relies on a bidirectional dependency network with lexical features and character n-grams, and the Bilty relies on a bidirectional RNN with concatenated word and character embeddings. We trained both taggers on WSJ (OntoNotes 4.0 version with some adaptions, see Appendix~\ref{app:wsj}) and report performance both in- and out-of-domain. For fair comparison, we did not use word clusters or word embeddings, but we otherwise used the best-performing models from the original papers\footnote{For Bilty, the best performance on WSJ was reported with 30 epochs and $\sigma$ = 0.3, so these values were used in our experiments as well.}.

\subsection{Evaluation}
The results, given in Table~\ref{tab:results}, are consistent with previous findings for POS taggers trained on WSJ, where ``tagging performance degrades on out-of-domain data'' \cite{gimpel-et-al-2011}. However, accuracy does not decrease by much for ESF compared to Twitter data, where accuracy dropped to 83.29\% \cite{derczynski-et-al-2013}. This is likely due to a greater degree of similarity between WSJ and ESF than WSJ and Twitter; unlike Twitter, ESF does not present the same affinity for ``conversational nature, […], lack of conventional orthography, and 140-character limit of each message'' \cite{gimpel-et-al-2011}. Performance mainly drops for out-of-vocabulary tokens, which makes up 12\% of tokens in ESF and 3\% of tokens in WSJ dev.

\begin{table}
    \begin{center}
    \def\arraystretch{.8}
    \begin{tabular}{llcc|llcc}
        \midrule
        \multicolumn{4}{c|}{\small{\textsc{Stanford}}} & \multicolumn{4}{c}{\small{\textsc{Bilty}}} \\
        
        $\hat{y}$ & $y$ & \# & \% & $\hat{y}$ & $y$ & \# & \% \\
        \midrule
        \textbf{\texttt{\small{NNP}}} & \textbf{\texttt{\small{NN}}} & \bf9 & \bf7.4 & \textbf{\texttt{\small{NNP}}} & \textbf{\texttt{\small{NN}}} & \bf10 & \bf8.9 \\
        \texttt{\small{NNP}} & \texttt{\small{NNPS}} & 7 & 5.8 & \texttt{\small{JJ}} & \texttt{\small{NN}} & 6 & 5.3 \\
        \texttt{\small{JJ}} & \texttt{\small{NN}} & 7 & 5.8 & \textbf{\texttt{\small{NNP}}} & \textbf{\texttt{\small{JJ}}} & \bf6 & \bf5.3 \\
        \textbf{\texttt{\small{NNP}}} & \textbf{\texttt{\small{JJ}}} & 7 & 5.8 & \texttt{\small{VBN}} & \texttt{\small{JJ}} & 5 & 4.4 \\
        \textbf{\texttt{\small{NN}}} & \textbf{\texttt{\small{NNP}}} & \bf5 & \bf4.1 & \texttt{\small{NNP}} & \texttt{\small{NNPS}} & 4 & 3.5 \\
        
        \textbf{\texttt{\small{NNP}}} & \textbf{\texttt{\small{NNS}}} & \bf4 & \bf3.3 & \textbf{\texttt{\small{NN}}} & \textbf{\texttt{\small{NNP}}} & \bf4 & \bf3.5 \\
        
        \texttt{\small{VBN}} & \texttt{\small{JJ}} & 4 & 3.3 & \textbf{\texttt{\small{NNP}}} & \textbf{\texttt{\small{NNS}}} & \bf4 & \bf3.5 \\
        
        & & & & \texttt{\small{NN}} & \texttt{\small{JJ}} & 4 & 3.5 \\
        
        & & & & \texttt{\small{VBD}} & \texttt{\small{VBN}} & 4 & 3.5 \\
        
        & & & & \texttt{\small{NN}} & \texttt{\small{NNS}} & 4 & 3.5 \\
        
        \midrule
    \end{tabular}
    \end{center}
    \caption{\label{tab:error-counts} Error counts of the most frequent error types (count $>$ 3). $\hat{y}$ are the predicted tags and $y$ are the gold tags. Proper noun errors are in bold.} 
\end{table}

\section{Error Analysis}

In addition to reporting the most frequent tag confusions (Table~\ref{tab:error-counts}), we qualitatively evaluated the output of each tagger and identified the following error categories:\\


\noindent \textbf{Notoriously Difficult Distinctions} 
\indent These are errors made on tokens that are notoriously difficult to classify, and where linguistic tests are needed to determine the part of speech in the given context \cite{santorini-macintyre-1995}. Examples include \textit{out}\texttt{/\{RB\textbar RP\textbar IN\}}, \textit{back}\texttt{/\{RB\textbar RP}\}, \textit{willing}\texttt{/\{VBG\textbar JJ\}}, and \textit{shattered}\texttt{/\{JJ\textbar VBN\}}. \\

\noindent \textbf{Tagging Conventions} 
\indent These are errors thought to be a consequence of WSJ not adhering to the guidelines by~\newcite{santorini-macintyre-1995}. An example is the token \textit{both}, which should be tagged as \texttt{CC} when it occurs with \textit{and}. In some cases, it had been tagged as \texttt{DT} instead.\\

\noindent \textbf{Encoding Errors}  
\indent These errors are a result of encoding issues and occur only with the Stanford tagger. Examples include dashes (―) being read as \textit{ÔÇö} or \textit{ÔÇò}, and trailing points (…) being read as \textit{ÔÇª}. 
The files had been encoded in UTF-8. \\

\noindent \textbf{Proper Noun Errors} 
\indent These are \texttt{NNP(S)} confusion errors. Examples are given in Table~\ref{tab:error-counts}. 
This category makes up 32.23\% of the errors made by the Stanford tagger and 27.43\% of the errors made by Bilty, though this number also includes any gold standard errors and cases where the taggers were expected to make errors. Consider, e.g., \textit{Restoring the Guardians}; being the name of a quest, it should be tagged as \textit{Restoring}\texttt{/NNP} \textit{the}\texttt{/DT} \textit{Guardians}\texttt{/NNPS}\footnote{Section 5.3 in \citet{santorini-macintyre-1995}.}, but both taggers have tagged \textit{Restoring} as \texttt{VBG}. \\

\noindent \textbf{Errors of Quantity}
\indent These are errors based on singular and plural tag confusion, e.g., \textit{folk} in \textit{for most folk (...)} being tagged as \texttt{NN} rather than \texttt{NNS}. \\

\noindent \textbf{Source Text Errors} 
\indent This category includes errors that likely stem from spelling errors or non-standard constructions in the source text, which complicates both the gold standard annotations and the machine annotations. Consider \textit{but good} in the sentence \textit{those guards sold you out but good}; it is unclear whether \textit{good} is meant as an adverb (\texttt{RB})\footnote{E.g., \textit{those guards sold you out \textbf{well\texttt{/RB}}}.} or as an adjective (\texttt{JJ})\footnote{E.g., \textit{those guards sold you out, but that's \textbf{good\texttt{/JJ}}}.}. \\

\noindent \textbf{Gold Standard Errors} 
\indent These are errors in the gold standard. For instance, \textit{nearby} was annotated as \texttt{RB} even though it was used as an \texttt{ADJ}, \textit{when} had been tagged as \texttt{IN} even though it was used in a temporal sense, and \textit{quite} had been tagged as \texttt{RB}, but it preceded a determiner, making it a \texttt{PDT}. \\ 

\noindent \textbf{Unexplainable Errors}
\indent These are errors where the cause for the erroneous prediction cannot be inferred from the context. Examples include \textit{sooner} being tagged as \texttt{RB} (adverb) instead of \texttt{RBR} (adverb in the comparative), and \textit{his} being tagged as \texttt{PRP\$} (possessive pronoun) even though it was used nominally (\texttt{PRP}).

\subsection{Errors on Known Tokens}
Many of the errors made by both taggers are notoriously difficult distinctions. Each tagger had also made additional, individual errors in this category, which emphasizes the difficulty of the task. The error analysis also revealed a few gold standard errors, as well as a number of proper noun errors; the latter is likely caused by incorrect capitalization in the source text. It should be noted, however, that in some cases, the correct part of speech (common noun vs. proper noun) can be considered disputable~\cite{manning2011part}.

The Stanford tagger made more mistakes in the unexplainable errors category, e.g., tagging \emph{will}\texttt{/MD} as \texttt{VBP} or \texttt{NN}, \emph{details}\texttt{/VBZ} and \emph{tasks}\texttt{/VBZ} as \texttt{NNS}, \emph{thaw}\texttt{/VB} and \emph{place}\texttt{/VB} as \texttt{NN}, and \emph{initiate}\texttt{/NN} as \texttt{VB} despite \emph{initiate} being preceded by \emph{an}\texttt{/DT}. Bilty did not make the same amount of unexplainable errors as the Stanford tagger, though it did tag the token \emph{have} incorrectly twice: once as \texttt{VBP} rather than the correct \texttt{VB}\footnote{\textit{(...) I do\texttt{/VBP} \textbf{have\texttt{/VB}} standards}.}, and once as \texttt{VB} rather than \texttt{VBP}\footnote{\textit{\textbf{Have\texttt{/VBP}} you heard about Goldenglow Meadery?}}, even though the tagger had correctly labeled the immediate surrounding tokens in both sentences. Bilty also had some issues with verbs that have the same form in the past tense and as a past participle, e.g., \emph{locked}, \emph{found} and \emph{called}, but only when they occured as participles without a main verb (e.g., when modifying nouns)\footnote{\textit{Tharn had a powerful weapon \textbf{called\texttt{/VBN}} the Staff of Chaos (...)}}. In these situations, Bilty incorrectly labeled the tokens as \texttt{VBD} instead of \texttt{VBN}. Bilty also incorrectly labeled \emph{impressed} as \texttt{VBN} rather than \texttt{JJ}, though this is likely due to inconsistent annotations in the training set.

\begin{table}
\setlength{\tabcolsep}{4pt}
\resizebox{1.\columnwidth}{!}{%
\begin{tabular}{l|rr|rr|rr}
\toprule
 & \multicolumn{2}{c}{Accuracy} & \multicolumn{2}{|c|}{NNP(S) Prec.} & \multicolumn{2}{c}{NNP(S) Rec.} \\
 & Orig. & Low. & Orig. & Low. & Orig. & Low. \\
 \midrule
Stanford & 94.19 & 89.40 & 89.50 & 78.05 & 94.23 & 46.15 \\
Bilty & 94.58 & 88.33 & 90.87 & 76.04 & 95.67 & 35.10 \\
\bottomrule
\end{tabular}}
\caption{\label{case experiment} Experiments with original casing (Orig.), and lowercased data (Low.) and  its influence on proper noun (\texttt{NNP+NNPS}) precision (Prec.) and recall (Rec.).}
\end{table}

\subsection{Errors on Unknown Tokens}
The majority of the errors made by both taggers on unknown tokens are proper noun errors. This is unsurprising, as the \emph{Elder Scrolls} universe introduces a number of fictional races and ethnicities that are rightfully capitalized, but often used as nouns (\texttt{NN}) or adjectives (\texttt{JJ}), e.g., \emph{Orsimer}, \emph{Nord}, and \emph{Khajiit}. Additionally, some tokens have been capitalized even when they do not represent proper nouns, leading both taggers to incorrectly tag them as \texttt{NNP}, e.g., \emph{Strength} and \emph{Endurance}. Both taggers also make errors of quantity, but they are not consistent in using, e.g., the suffix \textit{-s} as a determining factor; for instance, \emph{Nikulas} (the name of a character) was tagged as plural, but \emph{Scrolls} was tagged as singular. The taggers also struggled with terms from the \emph{Elder Scrolls} universe that have the same form in both singular and plural, e.g., \emph{Falmer}, \emph{Dunmer}, and \emph{Skaal}. Interestingly, both taggers make additional, unique proper noun errors where they each tag proper nouns as common nouns despite correct capitalization. The Stanford tagger, however, makes almost double the amount of these errors compared to the Bilty (7 vs. 4).

\subsection{Experiments with Letter Case}
Seeing as proper noun errors are a large source of errors for the unknown tokens, we attempted to increase robustness by lowercasing the data both for training and testing. This, however, turned out to considerably harm the detection of proper nouns for both taggers (Table~\ref{case experiment}), more so for Bilty, as indicated by the decrease in recall scores. Precision is not harmed as severely, but is still noticeably affected. This suggests that, even for datasets with inconsistent capitalization, capitalization aids the taggers more than it disrupts them.

\section{Conclusion}
This study presents a new, modest wiki dataset, Elder Scrolls Fandom (ESF), for qualitative cross-domain evaluation of POS taggers. Our analyses show that the Stanford tagger \cite{toutanova-et-al-2003} and Bilty~\cite{plank-et-al-2016}, a Bi-LSTM, both suffer a significant decrease in accuracy on unknown ESF tokens when trained on Wall Street Journal, one of the primary challenges being distinguishing proper nouns and other parts of speech. Lowercasing the data decreases accuracy even further, and negatively affects precision and recall on proper nouns; this shows that capitalization aids the taggers more than it disrupts them, even when inconsistent. Bilty performs slightly better than the Stanford tagger out-of-domain, the largest difference in accuracy being on unknown tokens: 78.37\% for Stanford vs. 80.41\% for Bilty.

\section*{Acknowledgments}
We would like to thank the NLPnorth and MaiNLP members as well as the anonymous reviewers for their feedback.

\bibliographystyle{acl_natbib}
\bibliography{references}

\begin{thebibliography}{8}
\expandafter\ifx\csname natexlab\endcsname\relax\def\natexlab#1{#1}\fi

\bibitem[{Bender and Friedman(2018)}]{bender-friedman-2018-data}
Emily~M. Bender and Batya Friedman. 2018.
\newblock Data statements for natural language processing: Toward mitigating
  system bias and enabling better science.
\newblock \emph{Transactions of the Association for Computational Linguistics},
  6:587--604.

\bibitem[{Derczynski et~al.(2013)Derczynski, Ritter, Clark, and
  Bontcheva}]{derczynski-et-al-2013}
Leon Derczynski, Alan Ritter, Sam Clark, and Kalina Bontcheva. 2013.
\newblock Twitter part-of-speech tagging for all: Overcoming sparse and noisy
  data.
\newblock In \emph{Proceedings of Recent Advances in Natural Language
  Processing}, pages 198--206.

\bibitem[{Gimpel et~al.(2011)Gimpel, Schneider, O'Connor, Das, Mills,
  Eisenstein, Heilman, Yogatama, Flanigan, and Smith}]{gimpel-et-al-2011}
Kevin Gimpel, Nathan Schneider, Brendan O'Connor, Dipanjan Das, Daniel Mills,
  Jacob Eisenstein, Michael Heilman, Dani Yogatama, Jeffrey Flanigan, and
  Noah~A. Smith. 2011.
\newblock Part-of-speech tagging for twitter: Annotation, features, and
  experiments.
\newblock In \emph{Proceedings of the 49th Annual Meeting of the Association
  for Computational Linguistics: shortpapers}, pages 42--47. Association for
  Computational Linguistics.

\bibitem[{Manning(2011)}]{manning2011part}
Christopher~D Manning. 2011.
\newblock Part-of-speech tagging from 97\% to 100\%: is it time for some
  linguistics?
\newblock In \emph{Computational Linguistics and Intelligent Text Processing:
  12th International Conference, CICLing 2011}, pages 171--189. Springer.

\bibitem[{Marcus et~al.(1993)Marcus, Santorini, and
  Marcinkiewicz}]{marcus-et-al-1993}
Mitchell~P. Marcus, Beatrice Santorini, and Mary~Ann Marcinkiewicz. 1993.
\newblock Building a large annotated corpus of english: The penn treebank.
\newblock \emph{Computational Linguistics}, 19(2):313--330.

\bibitem[{Plank et~al.(2016)Plank, Søgaard, and Goldberg}]{plank-et-al-2016}
Barbara Plank, Anders Søgaard, and Yoav Goldberg. 2016.
\newblock Multilingual part-of-speech tagging with bidirectional long
  short-term memory and auxiliary loss.
\newblock In \emph{Proceedings of the 54th Annual Meeting of the Association
  for Computational Linguistics}, pages 412--418. Association for Computational
  Linguistics.

\bibitem[{Santorini and MacIntyre(1995)}]{santorini-macintyre-1995}
Beatrice Santorini and Robert MacIntyre. 1995.
\newblock Part-of-speech tagging guidelines for the penn treebank project (3rd
  revision, 2nd printing).

\bibitem[{Toutanova et~al.(2003)Toutanova, Klein, Manning, and
  Singer}]{toutanova-et-al-2003}
Kristina Toutanova, Dan Klein, Christopher~D. Manning, and Yoram Singer. 2003.
\newblock Feature-rich part-of-speech tagging with a cyclic dependency network.
\newblock In \emph{Proceedings of HLT-NAACL 2003, Main Papers}, pages 173--180.

\end{thebibliography}
\clearpage

\appendix

\section{Adaptations to PTB}
\label{app:wsj}
\begin{figure}
    \centering
    \includegraphics[scale=0.60]{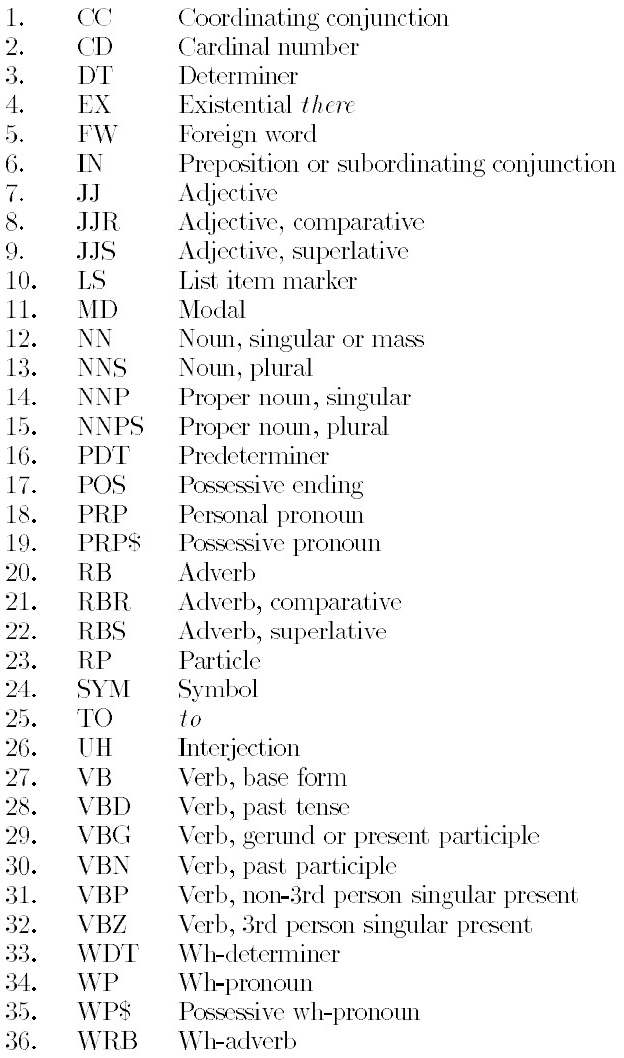}
    \caption{Word token tagset, borrowed from \newcite{santorini-macintyre-1995}. Slightly modified for whitespace and a page break.}
    \label{fig:fulltagset}
\end{figure}

\begin{figure}
    \centering
    \includegraphics[scale=0.65]{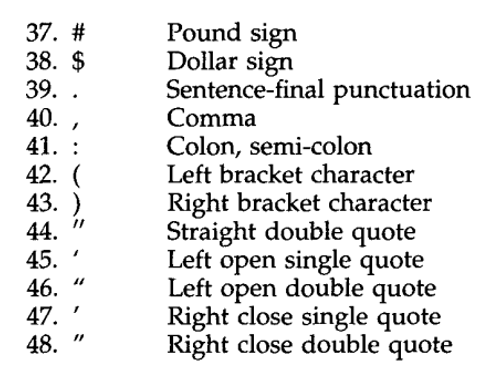}
    \caption{\label{fig:symboltagset} Original PTB symbol tagset, borrowed from \citet{marcus-et-al-1993}.}   
\end{figure}

We used the OntoNotes 4.0 release of the Wall Street Journal dataset (WSJ). For the ESF annotations, we followed the Penn Treebank (PTB) guidelines by \citet{santorini-macintyre-1995} for word tokens (see also Figure~\ref{fig:fulltagset}), but for symbol tokens, we found discrepancies between the PTB symbol tagset by \citet{marcus-et-al-1993} (see Figure~\ref{fig:symboltagset}), and the symbol tags used in WSJ. For this reason, we made the following adaptations to ensure comparability between ESF and WSJ:

\begin{itemize}
    \item We created a unified symbol tagset (Table \ref{esf symbol tagset}) for ESF and WSJ based on the tags used in OntoNotes 4.0.
    \item We adapted the WSJ data by replacing the bracket identifier tokens with the original tokens, e.g., \texttt{-LRB-} $\rightarrow$ \texttt{(}.
    \item We re-annotated all \textit{to} tokens in WSJ with the \texttt{TO} tag in accordance with \citeauthor{santorini-macintyre-1995}'s guidelines for PTB (\citeyear{santorini-macintyre-1995}).
\end{itemize}


\section{Data Statement}
\label{app:statement}
Following~\newcite{bender-friedman-2018-data}, the following outlines the data
statement for ESF:

\begin{enumerate}[label={\Alph*.}]
    \item \textsc{CURATION RATIONALE} \hspace{.1em} Collection of Elder Scrolls Fandom wiki data for fine-grained part-of-speech tagging and out-of-domain comparison with Wall Street Journal, including an error analysis.
    \item \textsc{LANGUAGE VARIETY} \hspace{.1em} Data was collected from the English wiki. The data contains more than one variety of English, but generally reflects US English (en-US).
    \item \textsc{SPEAKER DEMOGRAPHIC} \hspace{.1em} Unknown.
    \item \textsc{ANNOTATOR DEMOGRAPHIC} \hspace{.1em} One student with an academic background in English studies and one faculty member with 8 years of full-time experience in NLP research and UD annotation experience. Age range: 25-31. Gender: Female and male. White European. Native languages: Danish and Dutch. Socioeconomic status: Higher-education student and university faculty.
    \item \textsc{SPEECH SITUATION} \hspace{.1em} Written text with an option for continuous editing. Intended audience is expected to be fans of the \textit{Elder Scrolls} video game universe. Data has been generated between 2004 (when the Fandom platform was founded) and 2020 (the year of the database scraping the data was sourced from). The first \textit{Elder Scrolls} game predates the Fandom platform, making 2004 the lower bound. 
    \item \textsc{TEXT CHARACTERISTICS} \hspace{.1em} Randomly sampled sentences from articles or pages of encyclopedic nature. No formal editing, but continuous editing is possible. Topic is restricted to the \textit{Elder Scrolls} video game universe. Data includes quotes from the video games and character dialogue.
    \item \textsc{RECORDING QUALITY} \hspace{.1em} N/A.
    \item \textsc{OTHER} \hspace{.1em} The ESF dataset will be released under a CC BY-SA license\footnote{\url{https://creativecommons.org/licenses/by-sa/3.0/}} in accordance with the Fandom terms\footnote{\url{https://www.fandom.com/licensing}}.
    \item \textsc{PROVENANCE APPENDIX} \hspace{.1em} N/A.
\end{enumerate}

\section{ESF Tag Distribution}
The distribution of tags in ESF is given in Table~\ref{tab:tagdist}.

\begin{table}
\begin{center}
\begin{tabular}{c|c|m{3.9cm}}
\toprule
\textsc{Tag} & \textsc{Token} & \textsc{Name of Symbol(s)} \\ 
\midrule
\texttt{\small{,}} & , & Comma \\
\texttt{\small{.}} & . ? ! & Period, question mark, exclamation points \\
\texttt{\small{:}} & : ; - ... & Colon, semicolon, dash, trailing points \\
\texttt{\small{``}} & `` ` & Opening double and single quotes \\
\texttt{\small{''}} & " ' & Closing double and single quotes \\
\texttt{\small{\$}} & \$ \# & Dollar sign, pound sign \\
\texttt{\small{-LRB-}} & ( [ \{ & Left bracket, any kind \\
\texttt{\small{-RRB-}} & ) ] \} & Right bracket, any kind \\
\texttt{\small{SYM}} & / & Slash \\
\texttt{\small{NN}} & \% & Percent, percentage (both of which are nouns) \\
\texttt{\small{CC}} & \& & Ampersand (corresponds to the conjunction \emph{and}) \\
\bottomrule
\end{tabular}
\end{center}
\caption{\label{esf symbol tagset} ESF symbol tagset.}
\end{table}

\begin{table}
    \resizebox{1\columnwidth}{!}{
        \begin{tabular}{ccc|ccc}
            \toprule
            \textsc{Tag} & \textsc{Count} & \textsc{\%} & \textsc{Tag} & \textsc{Count} & \textsc{\%} \\
            \midrule
            \texttt{NN} & 239 & 11.47 & \texttt{PRP\$} & 23 & 1.1 \\
            \texttt{DT} & 236 & 11.32 & \texttt{``} & 21 & 1.01 \\
            \texttt{IN} & 222 & 10.65 & \texttt{''} & 21 & 1.01 \\
            \texttt{NNP} & 193 & 9.26 & \texttt{:} & 17 & 0.82 \\
            \texttt{.} & 118 & 5.66 & \texttt{NNPS} & 15 & 0.72 \\
            \texttt{PRP} & 97 & 4.65 & \texttt{POS} & 14 & 0.67 \\
            \texttt{,} & 91 & 4.37 & \texttt{WDT} & 10 & 0.48 \\
            \texttt{JJ} & 88 & 4.22 & \texttt{UH} & 8 & 0.38 \\
            \texttt{RB} & 88 & 4.22 & \texttt{WRB} & 8 & 0.38 \\
            \texttt{VB} & 80 & 3.84 & \texttt{CD} & 8 & 0.38 \\
            \texttt{VBZ} & 72 & 3.45 & \texttt{WP} & 6 & 0.29 \\
            \texttt{NNS} & 70 & 3.36 & \texttt{-RRB-} & 6 & 0.29 \\
            \texttt{CC} & 56 & 2.69 & \texttt{RP} & 5 & 0.24 \\
            \texttt{VBN} & 54 & 2.59 & \texttt{-LRB-} & 5 & 0.24 \\
            \texttt{TO} & 48 & 2.3 & \texttt{JJS} & 5 & 0.24 \\
            \texttt{VBD} & 44 & 2.11 & \texttt{JJR} & 4 & 0.19 \\
            \texttt{VBP} & 38 & 1.82 & \texttt{RBR} & 2 & 0.1 \\
            \texttt{VBG} & 38 & 1.82 & \texttt{RBS} & 1 & 0.05 \\
            \texttt{MD} & 32 & 1.54 & \texttt{PDT} & 1 & 0.05 \\
            \bottomrule
        \end{tabular}}
\caption{\label{tab:tagdist} ESF tag distribution.}
\end{table}

\end{document}